\pdfoutput=1
\documentclass[sigconf]{acmart}

\renewcommand\footnotetextcopyrightpermission[1]{} 
\usepackage{tikz}
\usepackage{pgfplots}
\usetikzlibrary{fit}
\usetikzlibrary{arrows,shapes}

\begin{document}

\title{Cross-lingual Extended Named Entity Classification of
  Wikipedia Articles}

\author{The Viet Bui}
\affiliation{FPT Technology Research Institute}
\affiliation{FPT University, Vietnam}
\email{vietbt6@fpt.com.vn}
\author{Phuong Le-Hong}
\affiliation{Vietnam National University, Hanoi, Vietnam}
\affiliation{FPT Technology Research Institute}
\email{phuonglh@vnu.edu.vn}

\begin{abstract}
  The FPT.AI team participated in the SHINRA2020-ML subtask of the
  NTCIR-15 SHINRA task.  This paper describes our method to solving
  the problem and discusses the official results. Our method focuses
  on learning cross-lingual representations, both on the word level
  and document level for page classification. We propose a three-stage
  approach including multilingual model pre-training, monolingual model
  fine-tuning and cross-lingual voting. Our system is able to achieve
  the best scores for 25 out of 30 languages; and its accuracy gaps to
  the best performing systems of the other five languages are
  relatively small.
\end{abstract}

\maketitle
\pagestyle{plain} 

\section*{Team Name}
FPT.AI

\section*{Subtask}
SHINRA2020-ML Shared Task


\section{Introduction}

SHINRA is a project to structure Wikipedia based on a pre-defined set
of attributes for given categories. The categories and the attributes
follow the definition of the Extended Named Entity
(ENE).\footnote{\url{https://ene-project.info}} Within this project, a
shared task called SHINRA2020-ML was proposed~\cite{Sekine:2020}. The
FPT.AI\footnote{\url{https://fpt.ai/}} team participated in this shared
task. This short paper describes our approach to solving the problem
and discusses our official results.

In this shared task, we are concerned with the problem of classifying
30 language Wikipedia entities in fine-grained categories, namely 219
categories defined in ENE. The FPT.AI team selects all the 30 target
languages to participate.  For each language, we run our system to
classify Wikipedia pages of that language and submit results for
evaluation.

Our method is inspired by an emerging
trend in learning general-purpose multilingual representations, which
can be applicable to many tasks, including text classification. Many
languages have similarities in syntax or vocabulary, and multiple
learning approaches that train on multiple languages while leveraging
the shared structure of the input space have begun to show good
results~\cite{Hu:2020}. We first develop a neural network
architecture which trains a multilingual model and then fine-tune this
model on language dependent datasets to obtain monolingual models. Our
neural network employs multilingual BERT~\cite{Devlin:2019} and a
special architecture for hierarchical multi-label classification,
which is specifically designed for maximizing the learning capacity
regarding the hierarchical structure of the labeled
data~\cite{Wehrmann:2018}. Finally, we propose a cross-lingual voting
technique to perform classification. 

Our method obtains good results on all languages. The FPT.AI system
achieves best scores for 25 out of 30 languages. For the other five
languages, its performance gaps to the best performing systems are
relatively small. 

The remainder of this paper is structured as
follows. Section~\ref{sec:method} describes our three-stage method and
the proposed neural network architecture. Section~\ref{sec:results} presents
experimental results. Section~\ref{sec:conclusion} concludes
the paper and suggests some directions for future work. 

\section{Method}
\label{sec:method}

Each sample in the dataset of a target language is a Wikipedia page
which contains some fields. The most important fields are 
\textit{opening text}, \textit{language}, \textit{title} and
\textit{text}. The \textit{text} field is the main content. In case
the page does not have the \textit{text} field, we use the
\textit{title} as the main content to perform classification.

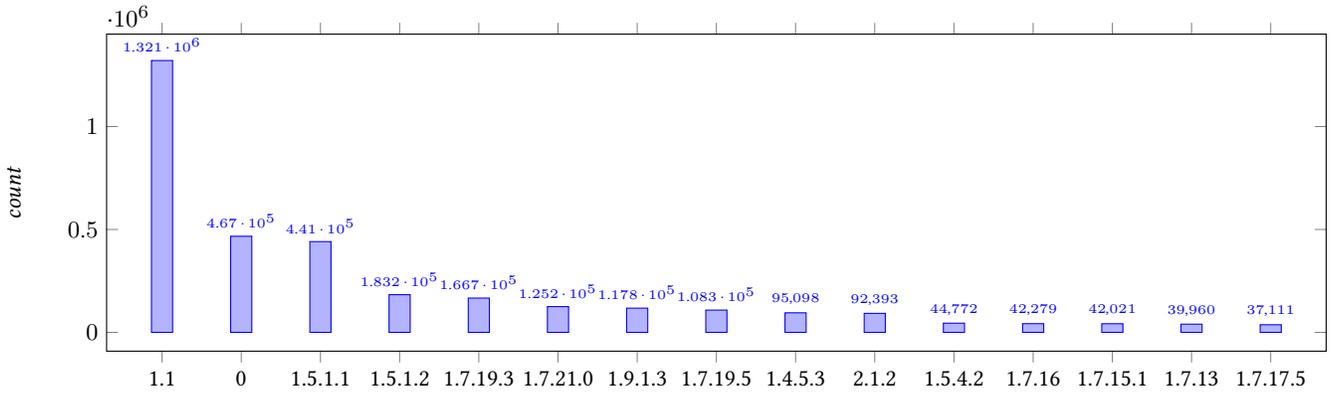
\begin{figure*}
  \centering
  \begin{tikzpicture}
    \begin{axis}[
      height=5.8cm,
      width=\textwidth,
      ybar,
      bar width=8,
      enlarge x limits=0.05,
      xlabel={\textit{}},
      ylabel={\textit{count}},
      symbolic x coords={1.1, 0, 1.5.1.1, 1.5.1.2, 1.7.19.3, 1.7.21.0,
      1.9.1.3, 1.7.19.5, 1.4.5.3, 2.1.2, 1.5.4.2, 1.7.16, 1.7.15.1,
      1.7.13, 1.7.17.5},
      xtick=data,
      every node near coord/.append style={font=\tiny},
      nodes near coords={\pgfmathprintnumber[precision=3]{\pgfplotspointmeta}},
      ]
      \addplot coordinates {(1.1,1320889) 
        (0,467016) (1.5.1.1,440989) (1.5.1.2,183176) (1.7.19.3,166701)
        (1.7.21.0,125235) (1.9.1.3,117789) (1.7.19.5,108291) (1.4.5.3,95098)
        (2.1.2,92393) (1.5.4.2,44772) (1.7.16,42279) (1.7.15.1,42021) (1.7.13,39960)
        (1.7.17.5,37111)};
    \end{axis}
  \end{tikzpicture}
  \caption{Top fifteen fine-grained ENE labels and their frequency}
  \label{fig:labels}  
\end{figure*}

Each category defined in ENEs has four levels, from a coarse-grained
type to a fine-grained type. More precisely, the first level $E_1$ has
5 labels; the second level $E_2$ has 25 labels, the third level $E_3$
has 94 labels, and the fourth level $E_4$ has 195 labels. The last
level contains the fine-grained types. This is essentially a
multi-label classification problem where each page need to be assigned
a subset of $E_4$ as its labels. Figure~\ref{fig:labels} shows the
partial histogram of the most frequent labels computed on all the target languages. We
see that the labels are highly imbalanced. Thus, we need to
deal with an imbalanced, hierarchical multi-label classification problem.

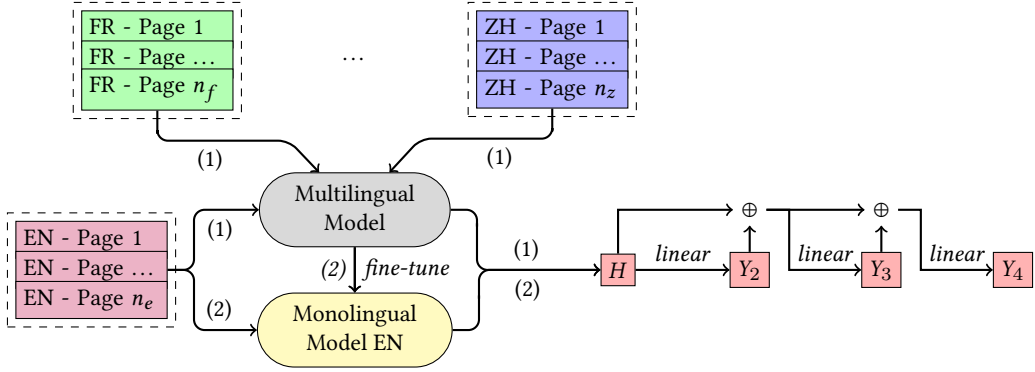
\begin{figure*}
  \centering
  \begin{tikzpicture}[x=1.75cm,y=0.8cm]
    \tikzstyle{every node} = [draw]
    \node[text width=1.8cm,fill=green!30] (f1) at (-1.5,4.5) {FR - Page 1};
    \node[text width=1.8cm,fill=green!30] (f2) at (-1.5,4.0) {FR - Page \dots};
    \node[text width=1.8cm,fill=green!30] (fN) at (-1.5,3.5) {FR - Page $n_f$};
    \node[rectangle,dashed,fill=none,fit={(f1) (f2) (fN)}] (fr) {};
    \node[draw=none,fill=none] (etc) at (0,4.0) {\dots};
    \node[text width=1.8cm,fill=blue!30] (z1) at (1.5,4.5) {ZH - Page 1};
    \node[text width=1.8cm,fill=blue!30] (z2) at (1.5,4.0) {ZH - Page \dots};
    \node[text width=1.8cm,fill=blue!30] (zN) at (1.5,3.5) {ZH - Page $n_z$};
    \node[rectangle,dashed,fill=none,fit={(z1) (z2) (zN)}] (zh) {};
    \node[text width=1.8cm,fill=purple!30] (e1) at (-2,1.0) {EN - Page 1};
    \node[text width=1.8cm,fill=purple!30] (e2) at (-2,0.5) {EN - Page \dots};
    \node[text width=1.8cm,fill=purple!30] (eN) at (-2,0.0) {EN - Page $n_e$};
    \node[rectangle,dashed,fill=none,fit={(e1) (e2) (eN)}] (en) {};

    \node[rounded rectangle,fill=gray!30] (multi) at (0,1.5) {\begin{tabular}{c}Multilingual\\Model\end{tabular}};
    \node[rounded rectangle,fill=yellow!30] (mono) at (0,-0.5)
    {\begin{tabular}{c}Monolingual\\Model EN\end{tabular}};

    \draw[->,thick] (fN.south) [rounded corners] -- ++(0,-0.5) |- ++(1.0,0) node[pos=0.7,below,draw=none] {(1)} -- (multi);
    \draw[->,thick] (zN.south) [rounded corners] -- ++(0,-0.5) |- ++(-1.0,0) node[pos=0.7,below,draw=none] {(1)} -- (multi);
    \draw[->,thick] (e2.east) [rounded corners] -- ++(0.2,0) |- node[pos=0.7,below,draw=none] {(1)} (multi);
    \draw[->,thick] (e2.east) [rounded corners] -- ++(0.2,0) |- node[pos=0.7,above,draw=none] {(2)} (mono);
    \draw[->,thick] (multi) -- node[pos=0.5,right,fill=none,draw=none] {\textit{fine-tune}} (mono);
    \draw[->,thick] (multi) -- node[pos=0.5,left,fill=none,draw=none] {\textit{(2)}} (mono);

    \node[fill=red!30] (H) at (2,0.5) {$H$};
    \node[fill=red!30] (Y2) at (3,0.5) {$Y_2$};
    \node[fill=red!30] (Y3) at (4,0.5) {$Y_3$};
    \node[fill=red!30] (Y4) at (5,0.5) {$Y_4$};
    \node[draw=none] (plus1) at (3,1.5) {$\oplus$};
    \node[draw=none] (plus2) at (4,1.5) {$\oplus$};
    \draw[->,thick] (multi.east) [rounded corners] -- ++(0.2,0) |- ++(0,-0.4) |- node[pos=0.7,above,draw=none] {(1)} (H.west);
    \draw[->,thick] (mono.east) [rounded corners] -- ++(0.2,0) |- ++(0,0.4) |- node[pos=0.7,below,draw=none] {(2)} (H.west);

    \draw[->,thick] (H) |- (plus1);
    \draw[->,thick] (Y2) -- (plus1);
    \draw[->,thick] (Y3) -- (plus2);
    \draw[->,thick] (plus1) -- (plus2);
    \draw[->,thick] (plus2) -- ++(0.3,0) |- node[pos=0.75,above,draw=none] {\textit{linear}} (Y4);
    \draw[->,thick] (H) -- node[pos=0.5,above,draw=none] {\textit{linear}} (Y2);
    \draw[->,thick] (plus1) -- ++(0.3,0) |- node[pos=0.75,above,draw=none] {\textit{linear}} (Y3);
  \end{tikzpicture}
  \caption{Cross-lingual named entity classification architecture} \label{fig:models}
\end{figure*}

We propose a three-stage method to tackle this problem. In the first
stage, we train a multilingual model for all the 30 languages using
BERT~\cite{Devlin:2019}. In the second stage, we fine-tune that model for each
language, and train monolingual models using the same BERT
architecture. Finally, in the third stage, we propose a simple voting
method to perform classification. Figure~\ref{fig:models} gives a high-level
overview of the first two stages. The subsequent subsections describe
these stages in detail. 

\subsection{Multilingual Model}

Let $P$ be an input Wikipedia page and $X$ be its main content, which
can be represented by a list of tokens $X = (x_1, x_2, \dots,
x_n)$. We need to find its output label set $Y$, which is a subset of
$E_1 \cup E_2 \cup E_3 \cup E_4$. This subset contains all the
terminal category in the ENE categories. We 
use SentencePiece, a language-independent subword tokenizer and
detokenizer~\cite{Kudo:2018} to tokenize $X$ into subwords using a
vocabulary $V$ of the BERT multilingual base cased model. The
multilingual BERT model was pretrained on the top 104 languages with
the largest Wikipedia using a masked language modeling (MLM)
objective~\cite{Devlin:2019}.

Let
$Z = \left( d_{[\text{CLS}]}, d_{s_1}, d_{s_2}, \dots, d_{s_t},
  d_{[\text{SEP}]} \right)$ be the corresponding index sequence of
$S$, where $d_k$ is the index of token $k$ in $V$ and
$t = \min\{m, l-2\}$. Here, $l$ is a hyper-parameter specifying the
max sequence length. These inputs are then fed into the multilingual
BERT (mBERT) model, and for each page $P$, we obtain its output vector
of $d_h$ dimensions:  
\begin{equation*}
  H = \text{mBERT}(Z)_{[\text{CLS}]} \in \mathbb R^{d_h}.
\end{equation*}

In order to make our model capable of labeling pages as belonging to
one or multiple paths in the hierarchy, we integrate HMCN, a
neural network architecture for hierarchical multi-label
classification~\cite{Wehrmann:2018}. HMCN is capable of simultaneously
optimizing local and global loss functions for discovering local
hierarchical relationships and global information from the
entire class hierarchy while penalizing hierarchical violations. This
method has achieved the state-of-the-art for hierarchical multi-class
classification.

In this work, we use the feed-forward (HMCN-F) architecture which is
specifically designed for maximizing the learning capacity regarding
the hierarchical structure of the labeled data. Let
$Y_i \in \mathbb R^{d_{E_i}}$ is the label information of $E_i$, where
$d_{E_i}$ is the number of labels in $E_i$, with $i \geq 2$. More
precisely:
\begin{align*}
  Y_2 &= \phi(H, d_{E_2})\\
  Y_3 &= \phi(H \oplus Y_2, d_{E_3})\\
  Y_4 &= \phi(H \oplus Y_2 \oplus Y_3, d_{E_4}),
\end{align*}
where $\oplus$ is the concatenation operator, $\phi(x,d)$ is a linear
function taking an input $x$ and an output dimension $d$. We do not
use $E_1$ because its number of labels is too small and there is
little information we can learn from $E_1$. Let
$Y_i, i = 2, 3, 4$ is the ground-truth one-hot vector corresponding to
the predicted target vector $\hat Y_i$. For each sample, we define a
loss function to be optimized as follows:
\begin{equation*}
  J(\theta) = \sum_{i=2}^4 L_i(\hat Y_i, Y_i, W_i),
\end{equation*}
where
\begin{equation*}
  L_i = -\frac{1}{M} \sum_{j=0}^M w_j \left[ y_j \log(\sigma(\hat y_j)) + (1 - y_j) \log(1 - \sigma(\hat y_j))\right],
\end{equation*}
where $\sigma(\cdot)$ is the sigmoid function
\begin{equation*}
 \sigma(z) = 1/(1 + \exp(-z)) 
\end{equation*}
and $w_j$ are weights which specify the importance of the $j$-th label
in the label vector $Y$. The weight vector
$w_i \in \mathbb R^{d_{E_i}}$ at level $i$ is computed as
\begin{equation*}
  w_i = \min \left( \frac{\vec k_i}{\vec c_i}, \mathbf 1 \right).
\end{equation*}
Here $\vec c_i$ is the frequency vector of all labels in $E_i$, $\vec k_i$ is
the element-wise average of $\vec c_i$ and $\mathbf 1$ is the unit vector of dimension
$d_{E_i}$.

\subsection{Monolingual Models}

After the first stage of training the multilingual model, in the
second stage, we fine-tune that model on each language, resulting in 
monolingual models. The monolingual models make use of the same neural
network architecture of the multilingual model as described in the
previous subsection.

In the prediction step, for each page $P$ in a given language, we
compute its prediction vector at the most fine-grained level $\hat
Y_4$ by passing the main content to the corresponding monolingual
model. This vector is then 
 passed to the sigmoid function $\hat Y = \sigma(\hat
Y_4)$, which squashes $\hat Y_4$ into a real-valued vector whose
elements are all in the $[0, 1]$ range. Finally, we pick the most
probably correct labels for $P$ by using a threshold $\epsilon$, that is the $l$-th label will be assigned to $P$ if
  $\hat Y_l \geq \epsilon$.

\subsection{Voting}

In this shared task, one is asked to classify each Wikipedia page into
ENE types where a page may be linked across different languages. Different
language-dependent pages share the same page identifier if they are
linked together by the interlanguage links. Thus, they should be
classified into the same hierarchical types. We leverage this
important property to boost the classification accuracy by using a
simple voting method as follows.

Let $P$ be a page and all its linked pages in $K$ other languages are
$P_1, P_2, \dots, P_K$. The monolingual models classify these pages
independently and the obtained result is $K$ lists of predicted
labels. These label lists are then flattened and their frequency
ratios $c_1, c_2, \dots, c_k$ are counted, where $k$ is the number of different
predicted 
labels for this page. Finally, all labels whose frequency is above the
average value will be chosen as the predicted label set for $P$. That
is, the $l$-th label will be chosen if
\begin{equation*}
  c_l \geq \frac{1}{k} \sum_{j=1}^k c_j.
\end{equation*}

\section{Results}
\label{sec:results}

\subsection{Datasets}

The organizer of this shared task provides the training data for 30
languages, created by the categorized Japanese Wikipedia of 920K pages
and Wikipedia language links for 30 languages.  The training data for
each target language may be a little bit noisy. For example, out of
2,263K German Wikipedia pages, 275K pages have a language link from
Japanese Wikipedia, which will be used as the training data for
German. Table~\ref{tab:datasets} shows some statistics of the Wikipedia data
for 31 languages that are processed by our system.

\begin{table}
  \centering
  \caption{Statistics of Wikipedia in 31 languages} \label{tab:datasets}
  \begin{tabular}{l | r | r | r}
    \hline
    Language&Pages&Links from JP&Ratio\\ \hline
    \hline 
    English (en)&5,790,377&439.354&7.6\\ \hline
    Spanish (es)&1,500,013&257,835&17.2\\ \hline
    French (fr)&2,074,648&318,828&15.4\\ \hline
    German (de)&2,262,582&274,732&12.1\\ \hline
    Chinese (zh)&1,041,039&267,107&25.7\\ \hline
    Russian (ru)&1,523,013&253,012&16.6\\ \hline
    Portuguese (pt)&1,014,832&217,896&21.5\\ \hline
    Italian (it)&1,496,975&270,295&18.1\\ \hline
    Arabic (ar)&661,205&73,054&11.0\\ \hline
    Japanese&1,136,222& --& --\\ \hline
    Indonesian (id)&451,336&115,643&25.6\\ \hline
    Turkish (tr)&321,937&111,592&34.7\\ \hline
    Dutch (nl)&1,955,483&199,983&10.2\\ \hline
    Polish (pl)&1,316,130&225,552&17.1\\ \hline
    Persian (fa)&660,487&169,053&25.6\\ \hline
    Swedish (sv)&3,759,167&180,948&4.8\\ \hline
    Vietnamese (vi)&1,200,157&116,280&9.7\\ \hline
    Korean (ko)&439,577&190,807&43.7\\ \hline
    Hebrew (he)&236,984&103,137&43.5\\ \hline
    Romanian (ro)&236,984&103,137&23.5\\ \hline
    Norwegian (no)&501,475&135,935&27.1\\ \hline
    Czech (cs)&420,195&135,935&25.1\\ \hline
    Ukrainian (uk)&420,195&135,935&20.5\\ \hline
    Hindi (hi)&129,141&30,547&23.6\\ \hline
    Finnish (fi)&450,537&144,750&32.1\\ \hline
    Hungarian (hu)&443,060&120,295&27.2\\ \hline
    Danish (da)&242,523&91,811&35.6\\ \hline
    Thai (th)&129,294&59,791&46.2\\ \hline
    Catalan (ca)&601,473&139,032&23.1\\ \hline
    Greek (el)&157,566&60,513&38.4\\ \hline
    Bulgarian (bg)&248,913&89,017&35.7 \\ \hline
  \end{tabular}
\end{table}

\subsection{Experimental Settings}

We use the following training details for our proposed neural network
architecture described in the previous section. The maximal sequence
length of each article is set to 512 tokens. The learning rate is
$2\times 10^{-5}$. The batch size is 45. The models are run in 100
epochs, using apex and BERT weight decay is set to 0. The rate of the
Adam optimizer is $10^{-8}$. The sigmoid threshold for label
assignment in monolingual models is set to 0.5. We use both CPU and
GPU devices for training and prediction. The CPU is an Intel Xeon (R)
E5-2699 v4 @2.20GHz. The GPU is a NVIDIA GeForce GTX 1080 Ti 11GB.

\begin{table}
  \centering
  \caption{Performance of the FPT.AI system} \label{tab:results}
  \begin{tabular}{l | l | c | c }
    \hline
    Lang.&Language&Regular&Late\\
    \hline
    \hline
    ar&Arabic&73.25&73.25\\ \hline
    bg&Bulgarian&\textbf{83.77}&83.28\\ \hline
    ca&Catalan, Valencian&52.55&\textcolor{blue}{81.10}\\ \hline
    cs&Czech&\textbf{84.47}&83.74\\ \hline
    da&Danish&\textbf{82.30}&81.74\\ \hline
    de&German&22.62&81.26\\ \hline
    el&Greek&\textbf{84.40}&84.10\\ \hline
    en&English&82.23&81.96\\ \hline
    es&Spanish, Castillian&80.60&80.60\\ \hline
    fa&Persian&\textbf{81.70}&81.52\\ \hline
    fi&Finnish&\textbf{83.62}&83.36\\ \hline
    fr&French&21.59&80.68\\ \hline
    he&Hebrew&\textbf{83.79}&\textcolor{blue}{84.21}\\ \hline
    hi&Hindi&\textbf{76.43}&75.65\\ \hline
    hu&Hungarian&\textbf{85.46}&84.78\\ \hline
    id&Indonesian&\textbf{81.93}&81.65\\ \hline
    it&Italian&26.55&\textcolor{blue}{82.81}\\ \hline
    ko&Korean&\textbf{83.67}&\textcolor{blue}{83.77}\\ \hline
    nl&Dutch, Flemish&\textbf{83.29}&83.17\\ \hline
    no&Norwegian&\textbf{80.53}&80.17\\ \hline
    pl&Polish&\textbf{84.53}&84.07\\ \hline
    pt&Portuguese&\textbf{83.23}&82.70\\ \hline
    ro&Romanian, Moldavian&\textbf{84.60}&84.60\\ \hline
    ru&Russian&\textbf{84.08}&83.44\\ \hline
    sv&Swedish&\textbf{83.18}&\textcolor{blue}{83.44} \\ \hline
    th&Thai&\textbf{81.26}&81.16\\ \hline
    tr&Turkish&\textbf{86.50}&86.03\\ \hline
    uk&Ukrainian&\textbf{83.12}&82.61\\ \hline
    vi&Vietnamese&\textbf{80.34}&\textcolor{blue}{80.42}\\ \hline
    zh&Chinese&\textbf{81.25}&80.60\\ \hline
  \end{tabular}
\end{table}

\subsection{Evaluation}

For each Wikipedia page, the system gives one or more predicted
fine-grained categories. A predicted category is considered correct if and
only if it is an exact match with the true category. The accuracy of
a system is evaluated using the micro average $F$ measure, which is
the harmonic mean of micro-averaged precision and micro-averaged
recall ratios. Note that the distribution of category in
the test data may differ from that of the training data or target
data. The organizer of the shared task keeps test datasets secret.

The FPT.AI system submitted the classification results for all the 30
target languages as regular submissions. Table~\ref{tab:results} shows
the $F$ measures of our system. The scores are extracted from the
official results published after the shared task by the
organizer~\cite{Sekine:2020}. Due to a technical problem, some of our
models did not perform well at the first submission (for example the
French model) and we tried to
re-submit the results a little bit after the deadline. The late
submission results are shown in the last column of
Table~\ref{tab:results}. The scores in bold font are the best scores
among participating systems. In the regular submission, the FPT.AI
system achieves the best scores for 23 languages. In the late
submission, our system is able to add two more languages into its
top-rank list, namely Catalan and Italian, achieving best scores for
25 out of 30 languages. In addition, this submission also pushes the
best scores further for four languages, including Hebrew, Korean,
Swedish, and Vietnamese.

\begin{table}
  \centering
  \caption{Performance gaps with other top systems} \label{tab:gaps}
  \begin{tabular}{l | c | c | c | l }
    \hline
    Language & FPT.AI Score & Best Score & $\Delta$ & Best Team\\
    \hline
    \hline
    Arabic&73.25&\textbf{76.27}&3.02&PribL\\ \hline
    German&81.26&\textbf{81.86}&0.60&Ousia\\ \hline
    French&80.68&\textbf{81.01}&0.33&Ousia\\ \hline
    English&82.23&\textbf{82.73}&0.50&Uomfj\\ \hline
    Spanish&80.60&\textbf{81.39}&0.79&Uomfj\\ \hline
  \end{tabular}
\end{table}

Our system is outperformed by other participating systems in five
languages, including Arabic, German, English, French and
Spanish. Table~\ref{tab:gaps} shows the score gaps between FPT.AI and
the best performing systems for each language. We see that the
performance gaps between our system with the top ones are relatively
small, except for the Arabic language.

\subsection{Ablation Analysis}

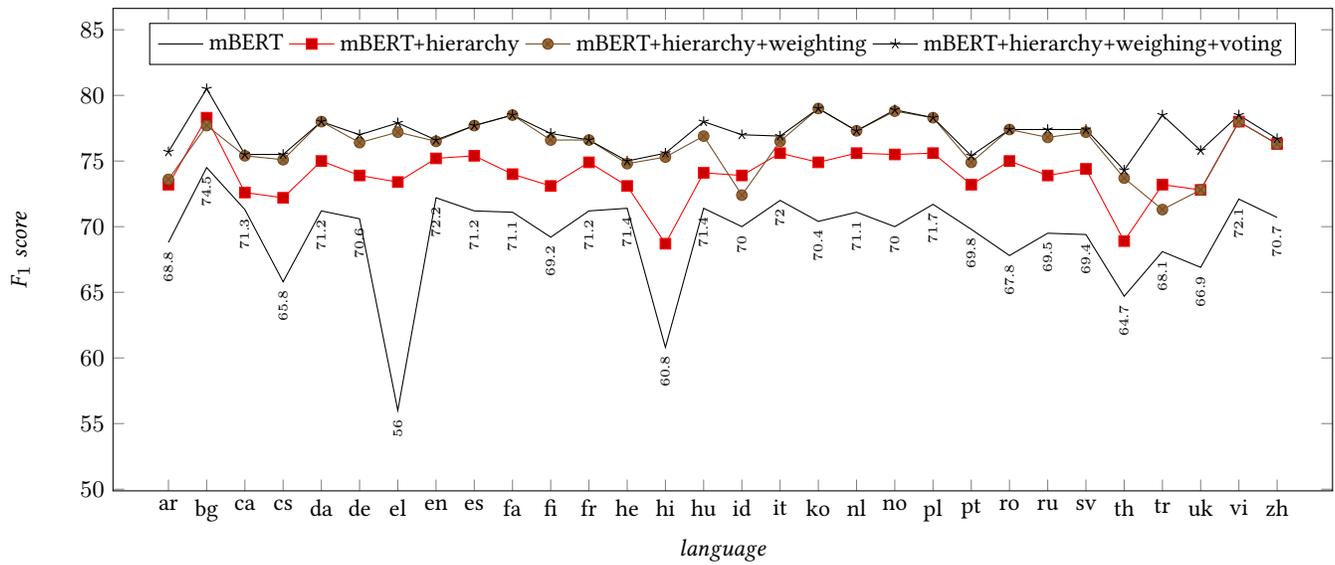
\begin{figure*}
  \begin{tikzpicture}
    \begin{axis}[
      height=8cm,
      width=\textwidth,
      enlarge x limits=0.05,
      enlarge y limits=0.25,
      legend style={at={(0.2,0.85)},anchor=south,legend columns=-1},
      xlabel={\textit{language}},
      ylabel={\textit{$F_1$ score}},
      symbolic x coords={ar, bg, ca, cs, da, de, el, en, es, fa, fi,
        fr, he, hi, hu, id, it, ko, nl, no, pl, pt, ro, ru, sv, th,
        tr, uk, vi, zh},
      xtick=data,
      legend pos = north east
      ]
      \addplot[
      nodes near coords,
      every node near coord/.append style={font=\tiny,rotate=90,anchor=east},
      nodes near coords={\pgfmathprintnumber[precision=3]{\pgfplotspointmeta}},
      nodes near coords align={vertical},
      ] coordinates {
        (ar, 68.8) (bg, 74.5) (ca, 71.3) (cs, 65.8) (da, 71.2) (de,
        70.6) (el, 56.0) (en, 72.2) (es, 71.2) (fa, 71.1) (fi, 69.2)
        (fr, 71.2) (he, 71.4) (hi, 60.8) (hu, 71.4) (id, 70.0) (it,
        72.0) (ko, 70.4) (nl, 71.1) (no, 70.0) (pl, 71.7) (pt, 69.8)
        (ro, 67.8) (ru, 69.5) (sv, 69.4) (th, 64.7) (tr, 68.1) (uk,
        66.9) (vi, 72.1) (zh, 70.7)};
      \addplot coordinates {
        (ar, 73.2) (bg, 78.3) (ca, 72.6) (cs, 72.2) (da, 75.0) (de,
        73.9) (el, 73.4) (en, 75.2) (es, 75.4) (fa, 74.0) (fi, 73.1)
        (fr, 74.9) (he, 73.1) (hi, 68.7) (hu, 74.1) (id, 73.9) (it,
        75.6) (ko, 74.9) (nl, 75.6) (no, 75.5) (pl, 75.6) (pt, 73.2)
        (ro, 75.0) (ru, 73.9) (sv, 74.4) (th, 68.9) (tr, 73.2) (uk,
        72.8) (vi, 78.0) (zh, 76.3)};
      \addplot coordinates {
        (ar, 73.6) (bg, 77.7) (ca, 75.4) (cs, 75.1) (da, 78.0) (de, 76.4) (el,
        77.2) (en, 76.5) (es, 77.7) (fa, 78.5) (fi, 76.6) (fr, 76.6) (he,
        74.8) (hi, 75.3) (hu, 76.9) (id, 72.4) (it, 76.5) (ko, 79.0) (nl,
        77.3) (no, 78.8) (pl, 78.3) (pt, 74.9) (ro, 77.4) (ru, 76.8) (sv,
        77.2) (th, 73.7) (tr, 71.3) (uk, 72.8) (vi, 78.0) (zh, 76.3)};
      \addplot coordinates {
        (ar, 75.7) (bg, 80.5) (ca, 75.5) (cs, 75.5) (da, 78.0) (de,
        77.0) (el, 77.9) (en, 76.6) (es, 77.7) (fa, 78.5) (fi, 77.1)
        (fr, 76.6) (he, 75.0) (hi, 75.6) (hu, 78.0) (id, 77.0) (it,
        76.9) (ko, 79.0) (nl, 77.3) (no, 78.9) (pl, 78.3) (pt, 75.4)
        (ro, 77.4) (ru, 77.4) (sv, 77.4) (th, 74.3) (tr, 78.5) (uk,
        75.8) (vi, 78.5) (zh, 76.7)}; 
      \legend{mBERT, mBERT+hierarchy, mBERT+hierarchy+weighting, mBERT+hierarchy+weighing+voting}
    \end{axis}
\end{tikzpicture}
  \caption{Performance of our model where parts of the method are
    systematically added.}\label{fig:ablation}
\end{figure*}

In this subsection, we perform an ablation analysis of our method,
where important components are systematically added. This experiment
helps investigate the contribution of the hierarchical classification
technique, the weighted loss function adjustment, and the voting
strategy to the final score. All the scores are evaluated on the
open leader-board datasets and directly copied from the leader-board
after each submission. Figure~\ref{fig:ablation} shows the result.

Adding the hierarchy-aware classification technique improves the plain
mBERT models by about 4.77\% of absolute score in average across 30
langauges. Using the weighted loss function improves further the
performance by about 2.10\% in average. Finally, the voting strategy
increases the performance further by about 0.9\%.

\section{Conclusion}
\label{sec:conclusion}

In this paper, we have described the method underlying the FPT.AI
system which participated in the SHINRA2020-ML shared task. The method
exploits a cross-lingual representations of Wikipedia pages in 30
languages before fine-tuning to specific monolingual models. The
method also relies on the strong transformers-based neural network
models mBERT and on a special treatment for hierarchical multi-label
classification which maximize learning capacity regarding the
hierarchical structure of the labeled data. Another important
technique in our method is a cross-lingual voting strategy which helps
select the most reliable categories for each page.

Our system is able to achieve best scores for 25 out of 30 languages;
and its performance gaps to the best performing systems of the other
five languages are relatively small.

Our method has been designed toward a general-purpose cross-lingual
representation and transfer learning, covering typologically diverse
languages. Although it is able to attain good performances on many
languages, especially on low resource ones, on high resource
languages it has not achieved the best accuracy compared to several other
strong participating systems. This suggests that there is room for
different fine-tuning methods for high-resource languages that may
improve further performance. This is an interesting research 
direction for our future work.

\section*{Acknowledgements}

We would like to thank the organizing committee of the SHINRA2020
project, notably Satoshi Sekine for introducing this interesting
shared tasks to us; Masako Nomoto and Koji Matsuda for supporting us
with technical issues that we encountered during the task.

\bibliographystyle{ACM-Reference-Format}

\bibliography{fptai}

\end{document}